\documentclass[letterpaper, 10 pt, conference]{ieeeconf}  % Comment this line out if you need a4paper

\IEEEoverridecommandlockouts                              % This command is only needed if 
\usepackage{epsfig} % for postscript graphics file
\usepackage{subcaption}
\usepackage{makecell}
\usepackage{afterpage}
\usepackage{amsmath}
\usepackage{cases}
\usepackage{cite}
\usepackage{url}

\title{\LARGE \bf
	CNN-SVO: Improving the Mapping in Semi-Direct Visual Odometry Using Single-Image Depth Prediction
}

\author{Shing Yan Loo$^{1,2}$, Ali Jahani Amiri$^{1}$, Syamsiah Mashohor$^{2}$, Sai Hong Tang$^{2}$ and Hong Zhang$^{1}$% <-this % stops a space
	\thanks{$^{1}$The authors are with Department of Computing Science, University of Alberta, Canada
		{\tt\small \{lsyan,jahaniam,hzhang\}@ualberta.ca}}%
	\thanks{$^{2}$The authors are with the Faculty of Engineering, Universiti Putra Malaysia, Malaysia
		{\tt\small \{syamsiah,saihong\}@upm.edu.my}}%
}

\begin{document}
	
	\maketitle
	\thispagestyle{empty}
	\pagestyle{empty}

	\begin{abstract}
		
	Reliable feature correspondence between frames is a critical step in visual odometry (VO) and visual simultaneous localization and mapping (V-SLAM) algorithms. In comparison with existing VO and V-SLAM algorithms, semi-direct visual odometry (SVO) has two main advantages that lead to state-of-the-art frame rate camera motion estimation: direct pixel correspondence and efficient implementation of probabilistic mapping method. This paper improves the SVO mapping by initializing the mean and the variance of the depth at a feature location according to the depth prediction from a single-image depth prediction network. By significantly reducing the depth uncertainty of the initialized map point (i.e., small variance centred about the depth prediction), the benefits are twofold: reliable feature correspondence between views and fast convergence to the true depth in order to create new map points. We evaluate our method with two outdoor datasets: KITTI dataset and Oxford Robotcar dataset. The experimental results indicate that the improved SVO mapping results in increased robustness and camera tracking accuracy.
	
	% Accordingly, SVO is able to match features of low surface texture and achieves state-of-the-art frame rate camera motion estimation.
		
	\end{abstract}

	\section{Introduction}
	\begin{figure}
		\centering
		\includegraphics[scale=0.8]{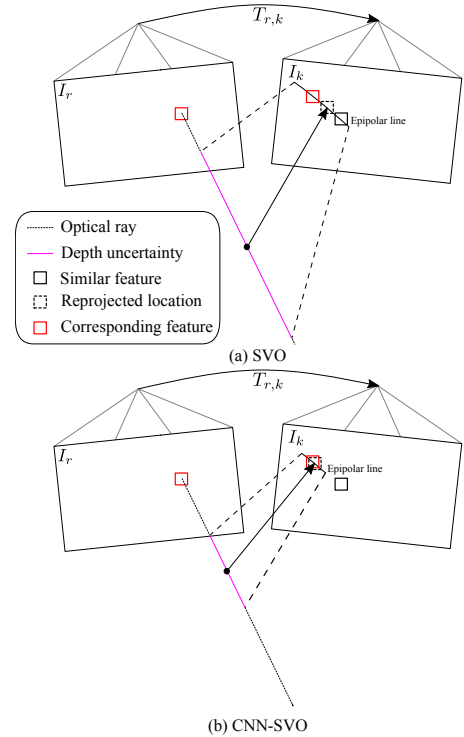}
		\caption{Proposed map point initialization strategy. Each initialized map point has a mean depth (black dot) and an interval in which the corresponding feature should lie, as shown by the magenta line. Note that larger depth uncertainty can allow the erroneous match to happen (as illustrated in (a) where the depth filter could converge to the "similar feature" rather than the "corresponding feature"). Our improved map point initialization method (see (b)) has lower depth uncertainty for identifying the corresponding feature }
		\label{fig:init_comparison}
	\end{figure}
	
	% Importance of visual odometry (motivation)
	Visual odometry (VO) and visual simultaneous localization and mapping (V-SLAM) have been actively researched and explored in the robotics field, including autonomous driving. As cameras become affordable and ubiquitous, being able to estimate camera poses reliably from an image sequence leads to important robotics applications, such as autonomous vehicle navigation.

	Matching features between the current frame and previous frames have been one of the most important steps in solving visual odometry (VO) and visual simultaneous localization and mapping (V-SLAM). There are two main feature matching methods: indirect method and direct method. Indirect methods \cite{RefWorks:doc:5ae22d97e4b09318b7184fd3,RefWorks:doc:58fa7ae1e4b07016f30e676e,RefWorks:doc:5ae22c3ee4b0a553e07595fb}  require feature extraction, feature description, and feature matching. These methods rely on matching the intermediate features (e.g., descriptors), and they perform poorly in images with weak gradients and textureless surfaces where descriptors are failed to be matched. Direct methods \cite{lsdslam,dso}, by contrast, do not need feature description, and they operate directly on pixel intensities; therefore, any arbitrary pixels (e.g., corners, edges, or the whole image) can be sampled and matched, resulting in reliable feature matching even in images with poor texture. However, matching pixels directly requires depths of the pixels to be recovered, and such matching is defined in the direct formulation that jointly optimizes structure and motion.
	
	Interestingly, semi-direct visual odometry (SVO) \cite{svo} is a hybrid method that combines the strength of direct and indirect methods for solving structure and motion, offering an efficient probabilistic mapping method to provide reliable map points for direct camera motion estimation. Unfortunately, one main limitation in SVO is that the map point is initialized with large depth uncertainty. Fig.~\ref{fig:init_comparison}(a) shows that the initialization of a map point with large depth uncertainty by SVO can lead to erroneous feature correspondence due to the large search range along the epipolar line.

	\begin{figure*}
		\centering
		\includegraphics[scale=1]{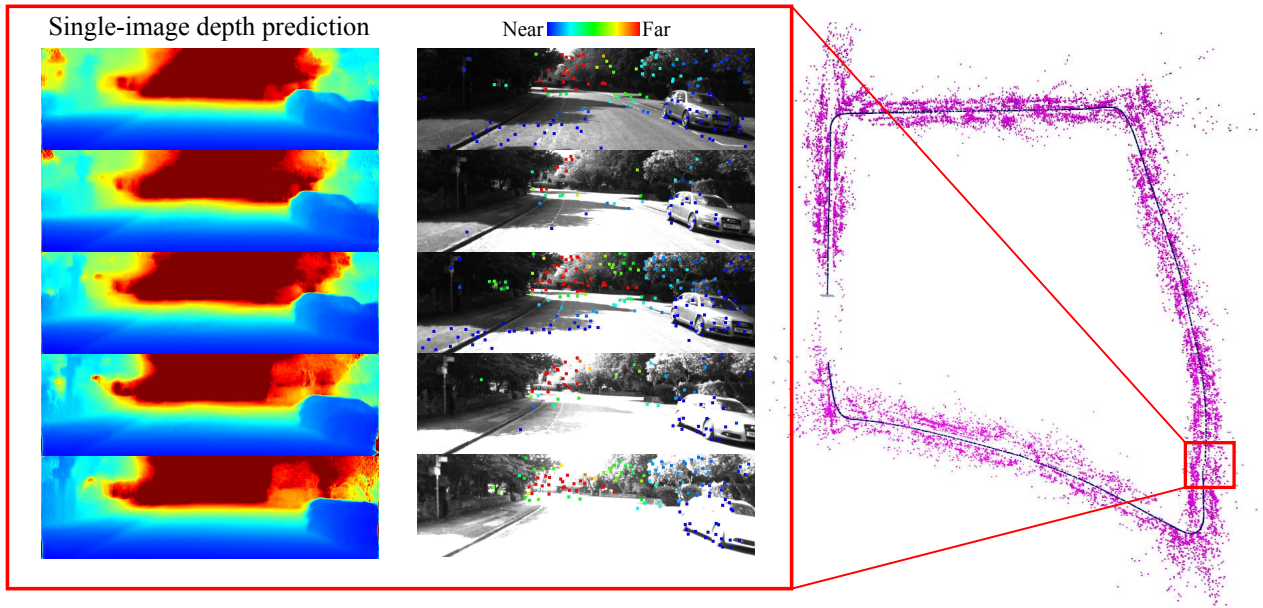}
		\caption{CNN-SVO: Camera motion estimation in the HDR environment. (Left) The single-image depth prediction model demonstrates the illumination invariance property in estimating depth maps, and the colour-coded reprojected map points on a sample sequence of five consecutive frames show the reprojected map points to those frames for camera motion estimation (best viewed in colour). Note that CNN-SVO only predicts depth maps for the keyframes. (Right) Camera trajectory (depicted with line) and map points in magenta generated by CNN-SVO\protect\footnotemark }
		\label{fig:cnnvo_overview}
	\end{figure*}
	
	In this paper, we propose to initialize new map points with depth prior from a single-image depth prediction neural network \cite{monodepth} (i.e., small variance centred about the predicted depth), such that the uncertainty for identifying the corresponding features is vastly reduced (see Fig.~\ref{fig:init_comparison}(b)). Because of the robust feature correspondence and small depth uncertainty, the map point is likely to converge to its true depth quickly. Overall, the improved SVO mapping, we refer to as CNN-SVO, is able to handle challenging lighting condition, thanks to the illumination invariance property in estimating depth maps (see Fig.~\ref{fig:cnnvo_overview}).

	\footnotetext{\url{https://github.com/yan99033/CNN-SVO}}
	
	\section{Methods} \label{sec:_methods}
	
	In this section, we briefly cover the working principle of SVO in Section~\ref{sec:_prelim} for the sake of completeness. Next, we detail our improved initialization of the map points in Section~\ref{sec:_mapping}.
	
	\subsection{Review of the SVO algorithm} \label{sec:_prelim}
	
	First, we explain the terminology used in the rest of the paper. A feature is one of the 2D points in the image extracted from FAST corner detector \cite{fastercornerdetector}. To perform feature correspondence, a small image patch that is centred at the feature location is used to find its corresponding patch in the nearby view; therefore, we refer to feature correspondence as matching small image patches. A map point is a 3D point projected from a feature location whose depth is known. 
	
	SVO \cite{svo} contains two threads running in parallel: tracking thread and mapping thread. In the tracking thread, the camera pose of a new frame is obtained by minimizing the photometric residuals between the reference image patches (from which the map points are back-projected) and the image patches that are centred at the reprojected locations in the new frame. The optimization steps for obtaining camera pose through the inverse compositional formulation \cite{inversecompositional} can be found in \cite{svo}. Concurrently, the mapping thread creates new map points using two processes: initialization of new map points with large depth uncertainty and update of depth uncertainty of the map points with \textit{depth-filters}; consequently, a new map point is inserted in the map if the depth uncertainty of the map point is small. 
	
	Given the camera poses of two frames one of which is the keyframe, the depth of a feature can be obtained using the following two steps: finding the feature correspondence along the epipolar line in the non-keyframe, and then recover the depth via triangulation. Since the occurrence of outlier matching is inevitable, a \textit{depth-filter} is modeled as a two-dimensional distribution  \cite{svo2,VOGIATZIS2011434}: the first dimension describes the probability distribution of the depth, and the second dimension models the inlier probability. Therefore, given a set of depth measurements, a \textit{depth-filter} approximates the mean depth and the variance (the first dimension) of the feature and separates the outliers from the inliers (the second dimension). The depth uncertainty (i.e., approximated variance) of the feature is updated when there is a new depth measurement, and the \textit{depth-filter} is considered to have converged if the updated depth uncertainty is small. Then, the converged \textit{depth-filters} that contain the true depths are used to create new map points by back-projecting the points at those feature locations according to their true depth. In this paper, we are focusing on improving the mapping in SVO \cite{svo}; therefore, we assume that the poses of the images can be successfully recovered in the tracking thread.
	
	% When a feature is extracted from a keyframe, the depth of at the feature location is intially unknown; therefore the depth estimate is modeled with a probability distribution that has a mean depth and a variance. 

	\subsection{Improved initialization of map points in SVO mapping} \label{sec:_mapping}
	
	The effective implementation of \textit{depth-filter} in SVO mapping and the use of direct matching of pixels has enabled SVO to achieve high frame rate camera motion estimation. However, SVO mapping initializes new map points in a reference keyframe with large uncertainty and their mean depths are set to the average scene depth in the reference frame. While such an initialization strategy is reasonable for the scene with one dominant plane---e.g., the floor plane---where the dominant depth information exists, the large depth uncertainty has limited the capability of the mapping to determine the true depths of the map points for the scene in which the depths of the map points vary considerably. Particularly, large depth uncertainty introduces two problems: possible erroneous feature correspondence along the epipolar line in the nearby frames and a high number of depth measurements to converge to the true depth. 
	
\begin{figure}
		\centering
		\includegraphics[scale=0.85]{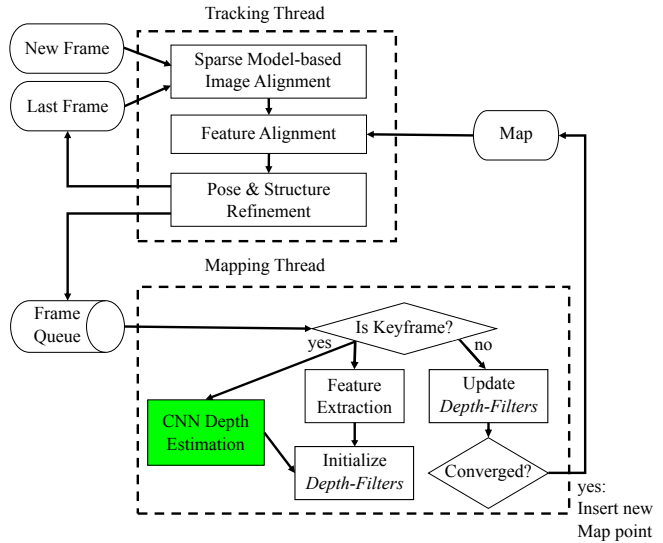}
		\caption{The CNN-SVO pipeline. Our work augments the SVO pipeline \cite{svo} with the CNN depth estimation module (marked in green) to improve the mapping in SVO }
		\label{fig:cnn_svo_pipeline}
	\end{figure}

	With single-image depth prediction as the prior knowledge of the scene geometry, our CNN-SVO able to obtain a much better estimate of the mean and a smaller initial variance of the \textit{depth-filter} than SVO to allow it to converge to the true depth of the map point. Fig.~\ref{fig:cnn_svo_pipeline} illustrates the CNN-SVO pipeline, in which we add the CNN depth estimation module (marked in green) to provide strong depth priors in the map points initialization process when a keyframe is selected---the initialization of \textit{depth-filters}. 
	
	Given a set of triangulated depth measurements, the goal of using \textit{depth-filter} is to separate the good measurements from the bad measurements: good measurements are normally distributed around the true depth, and bad measurements are uniformly distributed within an interval $ [\rho^{\tiny\textup{min}}_{i}, \rho^{\tiny\textup{max}}_{i}] $. Specifically given a set of triangulated inverse depth measurements $ \rho^{1}_{i}, \rho^{2}_{i},\cdots, \rho^{N}_{i} $ that correspond to the same feature, the measurement $ \rho_i^n $ is modeled in SVO using a \textit{Gaussian} + \textit{Uniform} mixture model:
	\begin{equation}
	p(\rho^{n}_{i} | \rho_{i}, \gamma_{i}) = \gamma_{i} \mathcal{N}(\rho^{n}_{i} | \rho_i, \tau^{2}_{i}) + (1-\gamma_{i}) \mathcal{U}(\rho^{n}_{i} | \rho^{\textup{min}}_{i}, \rho^{\textup{max}}_{i})
	\end{equation}
	where $ \rho_{i} $ is the true inverse depth, $ \tau^{2}_{i} $ the variance of the inverse depth, and $ \gamma_{i} $ the inlier ratio. Assuming the inverse depth measurements $ \rho^{1}_{i}, \rho^{2}_{i},\cdots, \rho^{N}_{i} $ are independent, \cite{svo2} shows that the approximation of the true inverse depth posterior can be computed incrementally by the product of a Gaussian distribution for the depth and a Beta distribution for the inlier ratio:
	\begin{equation}
	q(\rho_i,\gamma_i | a_n, b_n, \mu_n, \sigma_n^2) = Beta(\gamma_i | a_n, b_n) \mathcal{N}(\rho_i | \mu_n, \sigma_n^2)
	\end{equation}
	where $ a_n $ and $ b_n $ are the parameters in the Beta distribution, and $ \mu_n $ and $ \sigma_n^2 $ the mean and variance of the Gaussian depth estimate. The incremental Bayesian update step for $ a_n $, $ b_n $, $ \mu_n $, and $ \sigma_n^2 $ is described in detail in \cite{svo2,VOGIATZIS2011434}. Once $ \sigma_n^2 $ is lower than a threshold, the \textit{depth-filter} is converged to the true depth. 
	
\begin{figure}
		\centering
		\includegraphics[scale=1]{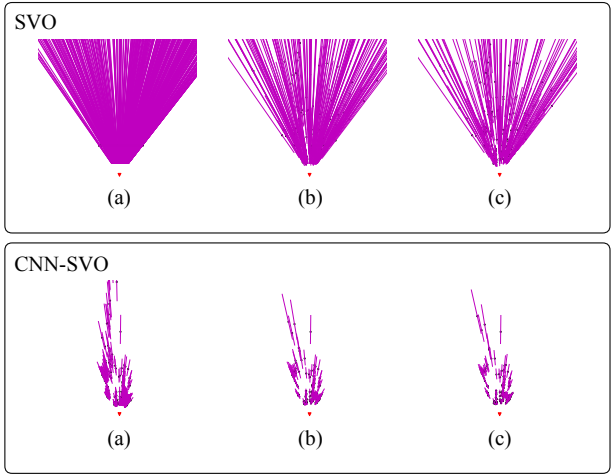}
		\caption{The improved mapping strategy is able to provide faster convergence of the map points. The length of the magenta line represents the depth uncertainty. (a) initialization of the depth filters where SVO uses a large interval to model the uncertainty of each initial map point whereas CNN-SVO uses a short interval; (b) depth estimates of the map points by the \textit{depth-filters} after three updates; (c) depth estimates of the map points by the \textit{depth-filters} after five updates }
		\label{fig:depthfiltering_kf}
	\end{figure}

	Hence, each \textit{depth-filter} is initialized with the following parameters: the mean of the inverse depth $ \mu_{n} $, the variance of the inverse depth $ \sigma_n^2 $, and the inlier ratio $ a_n $ and $ b_n $. Table~\ref{table:cnnsvo_params} compares the initialization of the parameters between SVO and CNN-SVO. The key difference is that CNN-SVO initializes the mean and the variance of the feature using learned scene depth instead of using the average and minimum scene depths in the reference keyframe. We empirically found that setting the depth variance to $ \frac{1}{(6d_{\tiny\textup{CNN}})^2} $ provides adequate room for noisy depth prediction to converge; we will be losing the absolute scale if the depth variance is large (e.g., replacing 6 with a higher number) by allowing more uncertainty in the measurement. Based on the initialized $ \mu_{n} $ and $ \sigma_n^2 $, a depth interval $ [\rho^{\tiny\textup{min}}_{i}, \rho^{\tiny\textup{max}}_{i}] $ can be defined by
	\begin{equation}
	\rho^{\tiny\textup{min}}_{i} = \mu_{n} + \sqrt{\sigma_n^2}
	\end{equation}
	\begin{subnumcases}{\rho^{\tiny\textup{max}}_{i}=}
	0.00000001, & \text{if}\ $\mu_{n} - \sqrt{\sigma_n^2} < 0$ 
	\\
	\mu_{n} - \sqrt{\sigma_n^2}, & \text{otherwise}
	\end{subnumcases}
	so that the corresponding feature can be found in the limited search range along the epipolar line in the nearby view (see Fig.~\ref{fig:init_comparison}). By obtaining strong depth prior from the single-image depth prediction network, the benefits are twofold: smaller uncertainty in identifying feature correspondence and faster map point convergence, as illustrated in Fig.~\ref{fig:depthfiltering_kf}.

	\begin{table}
		\begin{center}
			\caption{A comparison between SVO and CNN-SVO in the intialization of parameters. The parameters are defined by some prior knowledge of the scene, where $ d_{\tiny\textup{avg}} $ is the average scene depth in the reference keyframe, $ d_{\tiny\textup{CNN}} $ the depth prediction from the single-image depth prediction network, and  $ d_{\tiny\textup{min}} $ the minimum scene depth in the reference keyframe }
			\label{table:cnnsvo_params}
			\centering
			\renewcommand{\arraystretch}{1.5}
			\begin{tabular}{ccc}
				\Xhline{2\arrayrulewidth} % Make upper and lower hline thicker
				& SVO & CNN-SVO \\
				\hline
				$ \mu_{n} $ & $ \frac{1}{d_{\tiny\textup{avg}}} $ & $ \frac{1}{d_{\tiny\textup{CNN}}} $ \\
				$ \sigma_n^2 $ & $ \frac{1}{(6d_{\tiny\textup{min}})^2} $ & $ \frac{1}{(6d_{\tiny\textup{CNN}})^2} $ \\[1.5mm]
				\Xhline{2\arrayrulewidth} % Make upper and lower hline thicker
			\end{tabular}
		\end{center}
	\end{table}

	\section{Evaluation} \label{sec:_evaluation}
	We compare our method against the state-of-the-art direct and indirect methods, namely direct sparse odometry (DSO) \cite{dso}, semi-direct visual odometry (SVO) \cite{svo}, and ORB-SLAM without loop closure \cite{RefWorks:doc:591c741ce4b056205fce295a}. We use the absolute trajectory error (ATE) as the performance metric that has been used in the aforementioned papers. In addition, we indicate with 'X' for methods that are unable to complete the sequence due to lost tracking in the middle of the sequence (see Section~\ref{sec:_accuracy_evaluation}).
	
	To provide depth prediction in the initialization of map points in CNN-SVO, we adopt the Resnet50 variant of the encoder-decoder architecture from \cite{monodepth} that has already been trained on Cityscape dataset. Next, we fine-tune the network on stereo images in KITTI raw data excluding KITTI Odometry Sequence 00-10 using original settings in \cite{monodepth} for 50 epochs. To produce consistent structural information, even on overexposed or underexposed images, the brightness of the images has been randomly adjusted throughout the training, creating the effect of illumination variation. This consideration is useful for a neural network to handle high dynamic range (HDR) environments (see Fig.~\ref{fig:cnnvo_overview}).
	
	To design the system with real-time capability, we resize the images to $ 512\times 256 $ for depth map inference, and then we resize the depth map back to original shape for VO processing. While two separate threads have been designed to handle mapping and tracking, GPU is used to provide the depth maps for the keyframes.  The hardware is an Intel i7 processor\footnote{Intel i7-4790K, 4 cores, 4.0GHz, 32GB RAM} with NVidia GeForce GTX Titan X graphics card. 
	
	To scale the depth prediction for other datasets, the scaled depth $ d_{\textup{current}} $ can be obtained by the inferred depth $ d_{\textup{trained}} $ multiplied by the ratio of current focal length $ f_{\textup{current}} $ to trained focal length $ f_{\textup{trained}} $, that is:
	
	\begin{equation} \label{_eq_scale_depth}
	d_{\textup{current}} = \frac{f_{\textup{current}}}{f_{\textup{trained}}} d_{\textup{trained}}
	\end{equation}

	We use eleven KITTI Odometry sequences and nine Oxford Robotcar sequences for performance benchmarking. As for the images, we use the left camera from KITTI binocular stereo setup and the centre camera of the Bumblebee XB3 trinocular stereo setup from Oxford Robotcar. Both of the image streams are captured using global shutter cameras. Note that the ground truth poses from Oxford Robotcar dataset are not reliable for evaluation \cite{robotcar}, because of the poor and inconsistent GPS signals; we still use the ground truth for both quantitative and qualitative evaluation purposes. The frame rates are 10 frames per second (FPS) and 16 FPS for KITTI and Oxford Robotcar, respectively. To maintain the same aspect ratio that is used by the network input, the images in the Oxford Robotcar dataset have been cropped to 1248x376 throughout the evaluation process. We skip the first 200 frames for all the Oxford Robotcar sequences because of the extremely overexposed images at the beginning of the sequences. Since the network has not been trained on Oxford Robotcar dataset, we analyze the scale of the odometry relative to absolute scale for both datasets (see Section~\ref{sec:_scale_evaluation}).
	
	We set the maximum and the minimum number of tracked features in a frame to 200 and 100, respectively. Regarding the \textit{depth-filter}, we modify SVO to use 5 previous keyframes to increase the number of measurements in the \textit{depth-filters}. We also enable bundle adjustment during the evaluation process.
	
	\begin{figure*} 
		\centering
		\includegraphics[scale=1]{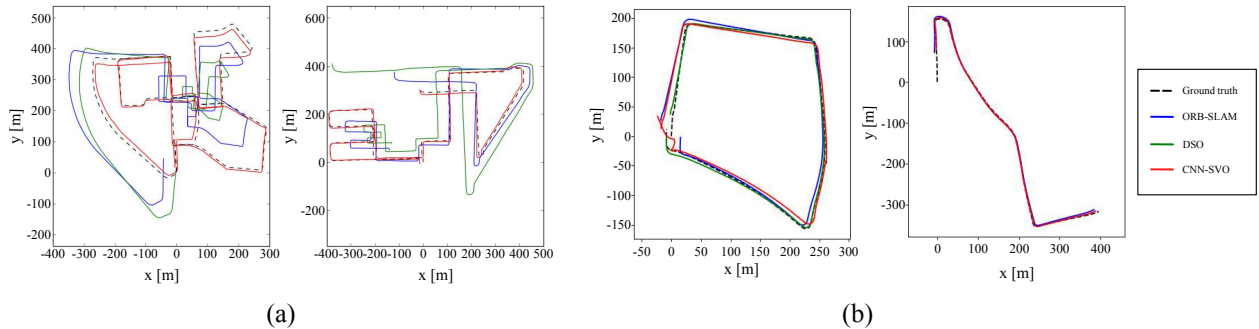}
		\caption{Qualitative comparison of camera trajectories produced by ORB-SLAM (without loop closure), DSO, and CNN-SVO. (a) KITTI Sequence 00 and 08; (b) Oxford Robotcar Sequence 2014-05-06-12-54-54 and 2014-06-25-16-22-15. SVO is not included in this figure because it is not able to complete the trajectory due to tracking and mapping failures }
		\label{fig:ate_combined}
	\end{figure*}
	
	\subsection{Accuracy evaluation} \label{sec:_accuracy_evaluation}	
	The ATEs of KITTI dataset and Oxford Robotcar dataset are collected with a median of 5 runs, and they are shown in Table~\ref{table:ate_kitti} and Table~\ref{table:ate_oxford}, respectively. Our system is able to track all the sequences except for KITTI Sequence 01, because of failure to match features accurately in the scene with repetitive structure. We also demonstrate that our competitors fail to track most of the Oxford Robotcar sequences, which contain severely overexposed images. While ORB-SLAM is able to track features in consistent lighting conditions, the vanished textural information in overexposed and underexposed images has resulted in failure to match feature in these HDR environments. The main reason of tracking failure in DSO is its inability of affine brightness modeling to handle severe brightness change in the sequences, and the problem has also been reported in stereo DSO \cite{stereodso}. Scale drift is also noticeable in the trajectories on KITTI dataset produced by DSO and ORB-SLAM (without loop closure). SVO is designed to perform well in a planar scene; therefore, it fails to identify corresponding features effectively in the outdoor scene, where depths of the features can vary considerably. We attribute the robust tracking of CNN-SVO to its ability to match features in consecutive frames with additional depth information, even when the images are overexposed or underexposed (see Fig.~\ref{fig:cnnvo_overview}). The qualitative comparison of the camera trajectories can be found in Fig.~\ref{fig:ate_combined} for KITTI dataset and Robotcar dataset, respectively. In Fig.~\ref{fig:ate_combined} (b), an S-like curve is produced by CNN-SVO near the end of trajectory in Sequence 2014-05-06-12-54-54, which is caused by a moving car in front of the camera. Since the network has not been trained on Oxford Robotcar sequences, the experimental results suggest generalization ability to the structurally similar scene.
	
		\begin{table}
		\begin{center}
			\caption{Absolute keyframe trajectory RMSE (in metre) on KITTI dataset}
			\label{table:ate_kitti}
			\centering
			\renewcommand{\arraystretch}{1.3}
			\begin{tabular}{ccccc}
				\Xhline{2\arrayrulewidth} % Make upper and lower hline thicker
				% \hline
				Sequence & SVO & CNN-SVO & DSO & \makecell{ORB-SLAM \\ (w/o loop closure)} \\
				\hline
				00 & X 		 & \textbf{17.5269}  & 113.1838        & 77.9502			\\
				01 & X 		 & X        		 & X	     	   & X					\\
				02 & X 	 	 & 50.5119  		 & 116.8108		   & \textbf{41.0064}	\\
				03 & X 		 & 3.4588            & 1.3943   	   & \textbf{1.0182}	\\
				04 & 58.3970 & 2.4414  		     & \textbf{0.422}  & 0.9302				\\
				05 & X 		 & \textbf{8.1513}   & 47.4605  	   & 40.3542			\\
				06 & X 		 & \textbf{11.5091}  & 55.6173  	   & 52.2282			\\
				07 & X 		 & \textbf{6.5141} 	 & 16.7192  	   & 16.546				\\
				08 & X 		 & \textbf{10.9755}  & 111.0832 	   & 51.6215			\\
				09 & X 		 & \textbf{10.6873}  & 52.2251  	   & 58.1742			\\
				10 & X 		 & \textbf{4.8354}   & 11.090   	   & 18.4765			\\
				\Xhline{2\arrayrulewidth} % Make upper and lower hline thicker
				% \hline
			\end{tabular}
		\end{center}
	\end{table}
	
	\begin{table}
		\begin{center}
			\caption{Absolute keyframe trajectory RMSE (in metre) on Oxford Robotcar dataset}
			\label{table:ate_oxford}
			\centering
			\renewcommand{\arraystretch}{1.3}
			\begin{tabular}{ccccc}
				% \hline
				\Xhline{2\arrayrulewidth} % Make upper and lower hline thicker
				Sequence & SVO & CNN-SVO & DSO & \makecell{ORB-SLAM \\ (w/o loop closure)} \\
				\hline
				\makecell{2014-05-06- \\ 12-54-54} & X & 8.657   			& \textbf{4.708} 	& 10.6596	\\
				\makecell{2014-05-06- \\ 13-09-52} & X & \textbf{9.1947}  	& X     			& X			\\
				\makecell{2014-05-06- \\ 13-14-58} & X & \textbf{10.1865} 	& X 				& X			\\
				\makecell{2014-05-06- \\ 13-17-51} & X & \textbf{8.26}    	& X	  				& X			\\
				\makecell{2014-05-14- \\ 13-46-12} & X & \textbf{13.7513} 	& X	  				& X			\\
				\makecell{2014-05-14- \\ 13-50-20} & X & \textbf{32.4199} 	& X  				& X			\\
				\makecell{2014-05-14- \\ 13-53-47} & X & \textbf{6.3017}  	& X  				& X			\\
				\makecell{2014-05-14- \\ 13-59-05} & X & 6.1515  			& \textbf{2.4532} 	& X			\\
				\makecell{2014-06-25- \\ 16-22-15} & X & \textbf{3.703}   	& X   				& 6.558		\\
				\Xhline{2\arrayrulewidth} % Make upper and lower hline thicker
				% \hline
			\end{tabular}
		\end{center}
	\end{table}
	
	\subsection{Runtime evaluation} \label{sec:_runtime_evaluation}
	Local BA (about 29 ms) and single-image depth prediction (about 37 ms) have been the most demanding processes in the pipeline, but both processes are only required when new keyframes are created. Despite the computational demand, we experimentally found that CNN-SVO runs faster at 16 FPS with Oxford Robotcar dataset than 10 FPS with KITTI dataset. This is due to the close distance between frames in high frame rate sequence, and hence lesser keyframes are selected relative to the total number of frames from the sequence. For this reason, real-time computation can be achieved.

	\subsection{Scale evaluation} \label{sec:_scale_evaluation}
	
	Since the network is trained on rectified stereo images with known baseline, we examine the scale of the odometry based on predicted depth from the network. Table~\ref{table:scale_kitti_oxford} (a) shows that the scale of the odometry is close to absolute scale in KITTI dataset because the training images are mostly from KITTI dataset. For Oxford Robotcar dataset, we scale the depth predictions using Eq.~\ref{_eq_scale_depth}, and the scale of the VO is between 0.9 and 0.97 (see Table~\ref{table:scale_kitti_oxford} (b)). We offer two possible explanations for the inconsistent odometry scale. First, as mentioned in the Oxford Robotcar dataset documentation, the provided ground truth poses are not accurate, and the reasons are as follows: inconsistent GPS signals and scale drift in the large-scale map (see Section III in \cite{robotcar}). Second, the single-image depth prediction network has not been trained on the images in the Oxford Robotcar dataset, so the recovery of absolute scale cannot be guaranteed.

	\begin{table}
		\begin{center}
			\caption{Scale relative to absolute scale in VO output from CNN-SVO}
			\label{table:scale_kitti_oxford}
			
			\begin{minipage}{.4\linewidth}
				\centering
				\subcaption{KITTI Dataset}
				\begin{tabular}{cc}
					% \hline
					\Xhline{2\arrayrulewidth} % Make line thicker
					Sequence & Scale \\
					\hline
					Sequence 00 &  0.9296 \\
					Sequence 01 & X \\
					Sequence 02 & 0.921 \\
					Sequence 03 & 1.0811 \\
					Sequence 04 & 1.1876 \\
					Sequence 05 & 0.9837 \\
					Sequence 06 & 0.9602 \\
					Sequence 07 & 1.0246 \\
					Sequence 08 & 1.0014 \\
					Sequence 09 & 1.043 \\
					Sequence 10 & 1.0512 \\
					% \hline
					\Xhline{2\arrayrulewidth}
				\end{tabular}
			\end{minipage}
			\hspace{0.4cm}
			\begin{minipage}{.4\linewidth}
				\centering
				\subcaption{Oxford Robotcar Dataset}
				\begin{tabular}{cc}
					% \hline
					\Xhline{2\arrayrulewidth}
					Sequence & Scale \\
					\hline
					2014-05-06-12-54-54 & 0.8953 \\
					2014-05-06-13-09-52 & 0.9321 \\
					2014-05-06-13-14-58 & 0.9172 \\
					2014-05-06-13-17-51 & 0.9399 \\
					2014-05-14-13-46-12 & 0.9103 \\
					2014-05-14-13-50-20 & 0.9737 \\
					2014-05-14-13-53-47 & 0.9427 \\
					2014-05-14-13-59-05 & 0.9473 \\
					2014-06-25-16-22-15 & 0.9236 \\
					% \hline
					\Xhline{2\arrayrulewidth}
				\end{tabular}
			\end{minipage}
		\end{center}
	\end{table}

	\section{Conclusion}
	In this paper, we have improved SVO mapping, called CNN-SVO, by initializing the map points with low uncertainty and the mean depth obtained from a single-image depth prediction neural network. The proposed method has two main advantages: (1) features can be matched effectively by limiting the search range along the epipolar line in nearby views, assuming the camera poses are known, and (2) the map points are initialized with lower depth uncertainty, therefore they are able to converge to their true depths faster. With the combination of single-image depth prediction and implementation of \textit{depth-filters}, CNN-SVO can perform mapping and estimate camera motion reliably. Thanks to the illumination invariance property in the single-image depth prediction network, depth maps produced from overexposed or underexposed images can still be used to facilitate feature correspondence between views, overcoming a key limitation of the original SVO.
	
	Nevertheless, there are still shortcomings we are planning to address in the future. First, the threshold of the map point uncertainty is increased to allow map points with larger uncertainty to be inserted for camera motion tracking. This is due to the limited observations of the corresponding features that can be found in the nearby frame as a consequence of limited frame rate. Hence, the increase in uncertainty threshold implicitly assumes accurate depth prediction from the single-image depth prediction network. Second, although the network is able to produce depth maps from overexposed images, it still could not produce useful depth map with blank image---i.e., completely overexposed image. Because blank images rarely occur in an extended period of time, we estimate the pose of the blank images with constant velocity model until new features can be extracted. Then local BA is applied to jointly correct the map points and camera poses. This problem can be mitigated using exposure compensation algorithm \cite{activeExposureHDR}. Lastly, we facilitate feature matching by limiting the search space of the corresponding feature along the epipolar line in nearby frames. This feature matching strategy does increase the tolerance of illumination change, but it does not solve the inherent problem of photometric constancy assumption in direct methods. Thus, incorporation of additional photometric calibration \cite{RefWorks:doc:5aa38ee1e4b01f401db446c2} can further improve the feature matching performance.
	
	\addtolength{\textheight}{-12cm}   % This command serves to balance the column lengths
	
	%%%%%%%%%%%%%%%%%%%%%%%%%%%%%%%%%%%%%%%%%%%d%%%%%%%%%%%%%%%%%%%%%%%%%%%%%%%%%%%%%

	%%%%%%%%%%%%%%%%%%%%%%%%%%%%%%%%%%%%%%%%%%%%%%%%%%%%%%%%%%%%%%%%%%%%%%%%%%%%%%%%

	%%%%%%%%%%%%%%%%%%%%%%%%%%%%%%%%%%%%%%%%%%%%%%%%%%%%%%%%%%%%%%%%%%%%%%%%%%%%%%%%
	%\section*{APPENDIX}
	
	
	%\section*{ACKNOWLEDGMENT}
	
	%The preferred spelling of the word ÒacknowledgmentÓ in America is without an ÒeÓ after the ÒgÓ. Avoid the stilted expression, ÒOne of us (R. B. G.) thanks . . .Ó  Instead, try ÒR. B. G. thanksÓ. Put sponsor acknowledgments in the unnumbered footnote on the first page.

	%%%%%%%%%%%%%%%%%%%%%%%%%%%%%%%%%%%%%%%%%%%%%%%%%%%%%%%%%%%%%%%%%%%%%%%%%%%%%%%%
	
	% References are important to the reader; therefore, each citation must be complete and correct. If at all possible, references should be commonly available publications.
	\bibliographystyle{IEEEtran}
	\bibliography{IEEEabrv,reference}
	
\end{document}